\documentclass[conference,a4paper]{lib/IEEEtran}
\IEEEoverridecommandlockouts
\usepackage{amsmath,amssymb,amsfonts}
\usepackage{algorithmic,algorithm}
\usepackage{graphicx}
\usepackage{textcomp}
\usepackage{xcolor}
\def\BibTeX{{\rm B\kern-.05em{\sc i\kern-.025em b}\kern-.08em
    T\kern-.1667em\lower.7ex\hbox{E}\kern-.125emX}}

\usepackage{times}
\usepackage{hyperref}
\usepackage{url}
\usepackage[utf8]{inputenc} 
\usepackage[T1]{fontenc}    
\usepackage{hyperref}       
\usepackage{url}            
\usepackage{booktabs}       
\usepackage{amsfonts}       
\usepackage{nicefrac}       
\usepackage{microtype}      
\usepackage{times}
\usepackage{graphicx}
\usepackage{subcaption} 
\usepackage{bbm}
\usepackage{amsmath}
\usepackage{amssymb}
\usepackage{color}
\usepackage{tabularx}
\usepackage{wrapfig}
\usepackage{caption}
\usepackage[capitalise]{cleveref}
\usepackage{xr}
\usepackage[numbers]{natbib}

\usepackage{amsmath,amsfonts,bm}

\def\eqref#1{equation~\ref{#1}}

\def\1{\bm{1}}

\DeclareMathAlphabet{\mathsfit}{\encodingdefault}{\sfdefault}{m}{sl}
\SetMathAlphabet{\mathsfit}{bold}{\encodingdefault}{\sfdefault}{bx}{n}

\newcommand{\ftheta}[1]{f_{\theta}(#1)}
\newcommand{\learnedLoss}{\mathcal{L}_\text{learned}}
\newcommand{\taskLoss}{\mathcal{L}_\mathcal{T}}
\newcommand{\fnoteleft}[1]{\footnote{\raggedright#1}}

\makeatletter
\newcommand*{\addFileDependency}[1]{
  \typeout{(#1)}
  \@addtofilelist{#1}
  \IfFileExists{#1}{}{\typeout{No file #1.}}
}
\makeatother
\newcommand*{\myexternaldocument}[1]{%
    \externaldocument{#1}%
    \addFileDependency{#1.tex}%
    \addFileDependency{#1.aux}%
}
\myexternaldocument{appendix}

\begin{document}

\title{Meta Learning via Learned Loss\\
\thanks{
\textsuperscript{*}Equal contributions.}
\thanks{
\textsuperscript{1}The authors are with the Max Planck Institute for Intelligent Systems, Tübingen, Germany}
\thanks{
\textsuperscript{2}The authors are with the Viterbi School of Engineering, 
University of Southern California,
Los Angeles, CA 90089}
\thanks{
\textsuperscript{3}The authors are with Facebook AI Research
}
\thanks{
\textsuperscript{4}The authors are with the Tandon School of Engineering, 
New York University,
Brooklyn, NY 11201}
}

\author{\IEEEauthorblockN{Sarah Bechtle\textsuperscript{*}\textsuperscript{1}}
\IEEEauthorblockA{
sbechtle@tuebingen.mpg.de }
\\
\IEEEauthorblockN{Edward Grefenstette\textsuperscript{3}}
\IEEEauthorblockA{
egrefen@fb.com  }
\and
\IEEEauthorblockN{Artem Molchanov\textsuperscript{*}\textsuperscript{2}}
\IEEEauthorblockA{
molchano@usc.edu}
\\
\IEEEauthorblockN{Ludovic Righetti\textsuperscript{1,4}}
\IEEEauthorblockA{
ludovic.righetti@nyu.edu}
\\
\IEEEauthorblockN{Franziska Meier\textsuperscript{3}}
\IEEEauthorblockA{
fmeier@fb.com  }
\and
\IEEEauthorblockN{Yevgen Chebotar\textsuperscript{*}\textsuperscript{2}}
\IEEEauthorblockA{
yevgen.chebotar@gmail.com }
\\
\IEEEauthorblockN{Gaurav Sukhatme\textsuperscript{2}}
\IEEEauthorblockA{
gaurav@usc.edu}
}

\maketitle
\begin{abstract}
    Typically, loss functions, regularization mechanisms and other important aspects of training parametric models are chosen heuristically from a limited set of options. In this paper, we take the first step towards automating this process, with the view of producing models which train faster and more robustly. Concretely, we present a meta-learning method for learning parametric loss functions that can generalize across different tasks and model architectures. We develop a pipeline for ``meta-training'' such loss functions, targeted at maximizing the performance of the model trained under them. The loss landscape produced by our learned losses significantly improves upon the original task-specific losses in both supervised and reinforcement learning tasks. Furthermore, we show that our meta-learning framework is flexible enough to incorporate additional information at \emph{meta-train} time. This information shapes the learned loss function such that the environment does not need to provide this information during \emph{meta-test} time. We make our code available at \url{https://sites.google.com/view/mlthree}
\end{abstract}


\section{Introduction}
\label{sec:intro}

Inspired by the remarkable capability of humans to quickly learn and adapt to new tasks, the concept of learning to learn, or \textit{meta-learning}, recently became popular within the machine learning community~\cite{l2l, rl2, maml}. 
We can classify \textit{learning to learn} methods into roughly two categories: approaches that learn representations that can generalize and are easily adaptable to new tasks~\cite{maml}, and approaches that learn how to optimize models~\cite{l2l, rl2}. 

\begin{figure}[t]
\begin{center}
\includegraphics[width=0.35\textwidth]{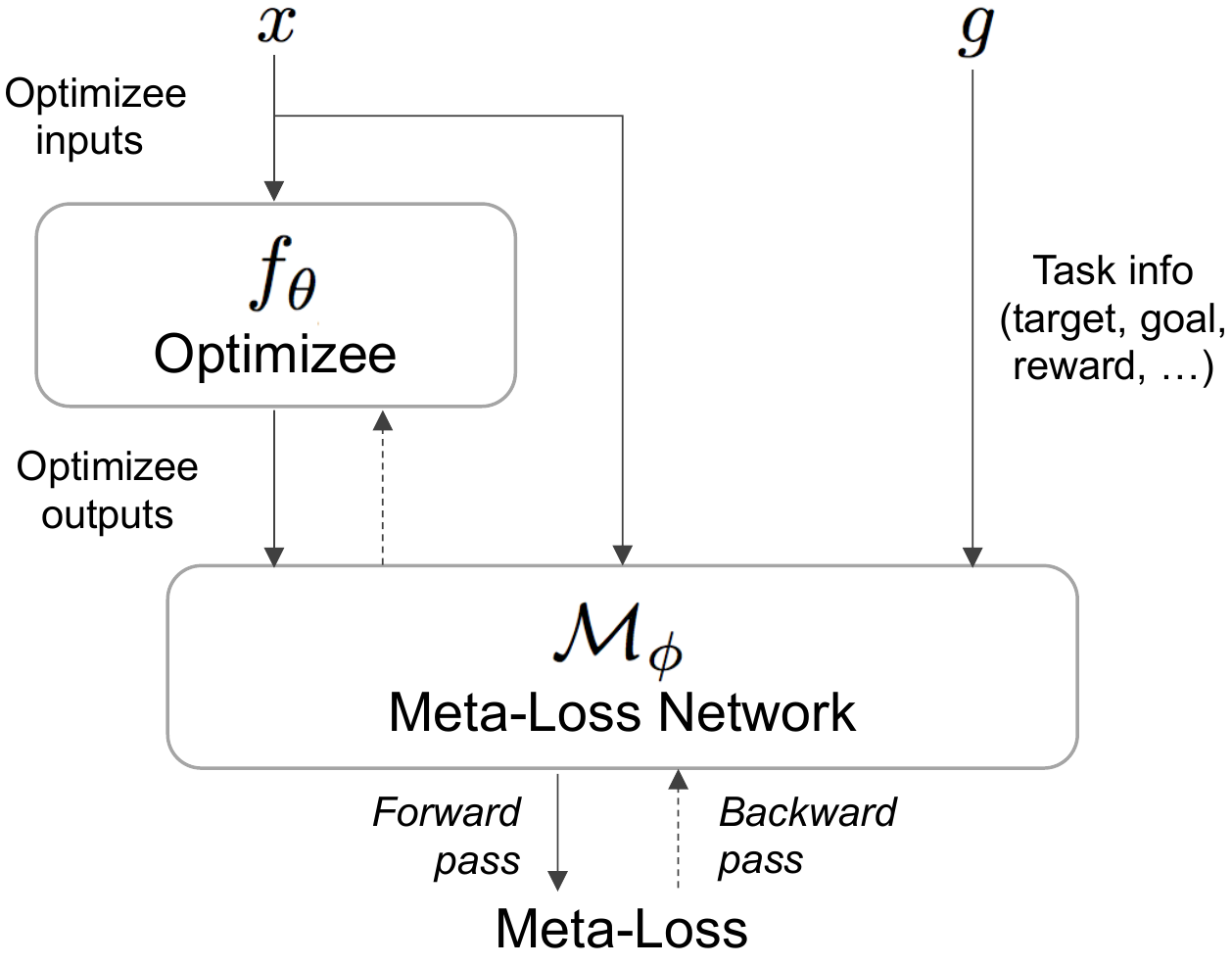}
 \caption{\small Framework overview: The learned meta-loss is used as a learning signal to optimize the optimizee $f_\theta$, which can be a regressor, a classifier or a control policy. } 
\label{fig:ml3_intro}
\end{center}
\vspace{-20pt}
\end{figure}

In this paper we investigate the second type of approach. 
We  propose a learning framework that is able to learn any parametric loss function---as long as its output is differentiable with respect to its parameters. Such learned functions can be used to efficiently optimize models for new tasks.

Specifically, the purpose of this work is to encode learning strategies into a parametric loss function, or a \textit{meta-loss}, which generalizes across multiple training contexts or tasks. 
Inspired by \textit{inverse reinforcement learning}~\cite{ng2000algorithms}, our work combines the \textit{learning to learn} paradigm of meta-learning with the generality of learning loss landscapes. 
We construct a unified, fully differentiable framework that can learn optimizee-independent loss functions to provide a strong learning signal for a variety of learning problems, such as classification, regression or reinforcement learning.
Our framework involves an inner and an outer optimization loops. 
In the inner loop, a model or an \textit{optimizee} is trained with gradient descent using the loss coming from our learned meta-loss function. 
\cref{fig:ml3_intro} shows the pipeline for updating the optimizee with the meta-loss. 
The outer loop optimizes the meta-loss function by minimizing a \emph{task-loss}, such as a standard regression or reinforcement-learning loss, that is induced by the updated optimizee.

The contributions of this work are as follows: 
i) we present a framework for learning adaptive, high-dimensional loss functions through back-propagation that create the loss landscapes for efficient optimization with gradient descent.
We show that our learned meta-loss functions improve over directly learning via the task-loss itself while maintaining the generality of the task-loss.
ii) We present several ways our framework can incorporate extra information that helps shape the loss landscapes at \emph{meta-train} time. 
This extra information can take on various forms, such as exploratory signals or expert demonstrations for RL tasks. 
After training the meta-loss function, the task-specific losses are no longer required since the training of optimizees can be performed entirely by using the meta-loss function alone, without requiring the extra information given at \emph{meta-train} time. 
In this way, our meta-loss can find more efficient ways to optimize the original task loss. 

We apply our meta-learning approach to a diverse set of problems demonstrating our framework's flexibility and generality.
The problems include regression problems, image classification, behavior cloning, model-based and model-free reinforcement learning. 
Our experiments include empirical evaluation for each of the aforementioned problems.
%
\section{Related Work}
\label{sec:related}
Meta-learning originates from the concept of learning to learn~\cite{Schmidhuber:87long, bengio:synaptic, ThrunP98}. 
Recently, there has been a wide interest in finding ways to improve learning speeds and generalization to new tasks through meta-learning. Let us consider gradient based learning approaches, that update the parameters of an \emph{optimizee} $f_\theta(x)$, with model parameters $\theta$ and inputs $x$ as follows:
\begin{align}
\vspace{-15pt}
    \theta_\text{new} = h_{\psi}(\theta, \nabla_\theta \mathcal{L}_\phi(y, f_\theta(x));
\vspace{-5pt}
\end{align}
where we take the gradient of a loss function $\mathcal{L}$, parametrized by $\phi$, with respect to the optimizee's parameters $\theta$ and use a gradient transform $h$, parametrized by $\psi$, to compute new model parameters $\theta_\text{new}$\fnoteleft{For simple gradient descent: $~~~~~~~~h(\theta, \nabla_\theta \mathcal{L}(y, f_\theta(x)) = \theta - \psi \nabla_\theta \mathcal{L}(y, f_\theta(x))$}. 
In this context, we can divide related work on meta-learning into learning model parameters $\theta$ that can be easily adapted to new tasks~\cite{maml, mendonca2019guided, gupta2018meta,yu2018one}, learning optimizer policies $h$ that transform parameters updates with respect to known loss or reward functions~\cite{maclaurin2015gradient, l2l, li2016learning, franceschi2017forward, meier2018online, rl2}, or learning loss/reward function representations $\phi$~\cite{sung2017learning, epg18, zou2019reward}. 
Alternatively, in unsupervised learning settings, meta-learning has been used to learn unsupervised rules that can be transferred between tasks~\cite{metz2018learning, cactus}.

Our framework falls into the category of learning loss landscapes. Similar to works by~\citet{sung2017learning} and~\citet{epg18}, we aim at learning loss function parameters $\phi$ that can be applied to various optimizee models, e.g. regressors, classifiers or agent policies. 
Our learned loss functions are independent of the model parameters $\theta$ that are to be optimized, thus they can be easily transferred to other optimizee models. This is in contrast to methods that meta-learn model-parameters $\theta$ directly~\cite[e.g.][]{maml, mendonca2019guided}, which are orthogonal and complementary to ours, where the learned representation $\theta$ cannot be separated from the original model of the optimizee. 
The idea of learning loss landscapes or reward functions in the reinforcement learning (RL) setting can be traced back to the field of inverse reinforcement learning~\cite[IRL]{ng2000algorithms, an04}. 
However, in contrast to IRL we do not require expert demonstrations (however we can incorporate them).
Instead we use task losses as a measure of the effectiveness of our loss function when using it to update an optimizee.

Closest to our method are the works on \textit{evolved policy gradients}~\cite{epg18}, \textit{teacher networks}~\cite{WuTXFQLL18}, \textit{meta-critics}~\cite{sung2017learning} and \textit{meta-gradient RL}~\cite{xusilver}.
In contrast to using an evolutionary approach~\cite[e.g.][]{epg18}, we design a differentiable framework and describe a way to optimize the loss function with gradient descent in both supervised and reinforcement learning settings. 
\citet{WuTXFQLL18} propose that instead of learning a differentiable loss function directly, a teacher network is trained to predict parameters of a manually designed loss function, whereas each new loss function class requires a new teacher network design and training. In~\citet{xusilver}, discount and bootstrapping parameters are learned online to optimize a task-specific meta-objective.
Our method does not require manual design of the loss function parameterization or choosing particular parameters that have to be optimized, as our loss functions are learned entirely from data. 
Finally, in work by~\citet{sung2017learning} a \textit{meta-critic} is learned to provide a task-conditional value function, used to train an actor policy. 
Although training a meta-critic in the supervised setting reduces to learning a loss function as in our work, in the reinforcement learning setting we show that it is possible to use learned loss functions to optimize policies directly with gradient descent. 

\vspace{-1pt}

\section{Meta-Learning via Learned Loss}
\label{sec:approach}
In this work, we aim to learn a loss function, which we call \textit{meta-loss}, that is subsequently used to train an \textit{optimizee}, e.g. a classifier, a regressor or a control policy. 
More concretely, we aim to learn a meta-loss function $\mathcal{M}_\phi$ with parameters $\phi$, that outputs the loss value $\learnedLoss$ which is used to train an optimizee $f_\theta$ with parameters $\theta$ via gradient descent:
\begin{align}
\vspace{-15pt}
\label{eq:pol_update}
\theta_\text{new} = \theta - \alpha \nabla_{\theta}\learnedLoss,\\ \text{ where } \learnedLoss = \mathcal{M}_\phi(y, f_{\theta}(x))
\vspace{-3pt}
\end{align}
where $y$ can be ground truth target information in supervised learning settings or goal and state information for reinforcement learning settings. 
In short, we aim to learn a loss function that can be used as depicted in Algorithm~\ref{algo:ml3_test_sup}. 
Towards this goal, we propose an algorithm to learn meta-loss function parameters $\phi$ via gradient descent. 

The key challenge is to derive a training signal for learning the loss parameters $\phi$. 
In the following, we describe our approach to addressing this challenge, which we call \textbf{M}eta-\textbf{L}earning via \textbf{L}earned \textbf{L}oss (ML$^3$).

\vspace{-0.1cm}
\subsection{\texorpdfstring{ML$^3$}{Lg} for Supervised Learning}\label{sec:ml3_supervised}
\vspace{-3pt}
We start with supervised learning settings, in which our framework aims at learning a meta-loss function $\mathcal{M}_\phi(y, f_\theta(x))$ that produces the loss value given the ground truth target $y$ and the predicted target $f_\theta(x)$. 
For clarity purposes we constrain the following presentation to learning a meta-loss network that produces the loss value for training a regressor $f_\theta$ via gradient descent, however the methodology trivially generalizes to classification tasks. 

Our meta-learning framework starts with randomly initialized model parameters $\theta$ and loss parameters $\phi$. 
The current loss parameters are then used to produce loss value $\learnedLoss = \mathcal{M}_\phi(y, f_\theta(x))$. 
To optimize model parameters $\theta$ we need to compute the gradient of the loss value with respect to $\theta$, $\nabla_\theta \mathcal{L} = \nabla_\theta \mathcal{M}_\phi(y, f_\theta(x))$. 
Using the chain rule, we can decompose the gradient computation into the gradient of the loss network with respect to predictions of model $f_\theta(x)$ times the gradient of model $f$ with respect to model parameters\footnote{Alternatively this gradient computation can be performed using automatic differentiation}, 
\begin{align}
\nabla_\theta \mathcal{M}_\phi(y, f_\theta(x)) = \nabla_{f} \mathcal{M}_\phi(y, f_\theta(x)) \nabla_\theta f_\theta(x). 
\end{align}
Once we have updated the model parameters \mbox{$\theta_\text{new} = \theta - \alpha \nabla_{\theta}\learnedLoss$} using the current meta-loss network parameters $\phi$, we want to measure how much learning progress has been made with loss-parameters $\phi$ and optimize $\phi$ via gradient descent.
Note, that the new model parameters $\theta_\text{new}$ are implicitly a function of loss-parameters $\phi$, because changing $\phi$ would lead to different $\theta_\text{new}$. 
In order to evaluate $\theta_\text{new}$, and through that loss-parameters $\phi$, we introduce the notion of a \emph{task-loss} during \emph{meta-train} time. 
For instance, we use the mean-squared-error (MSE) loss, which is typically used for regression tasks, as a task-loss 
$\taskLoss = (y - f_{\theta_\text{new}}(x))^2$.
We now optimize loss parameters $\phi$ by taking the gradient of $\taskLoss$ with respect to $\phi$ as follows\footnotemark[2]:
\begin{align}
\vspace{-3pt}
    & \nabla_{\phi} \taskLoss = \nabla_{f} \taskLoss \nabla_{\theta_\text{new}} f_{\theta_\text{new}} \nabla_{\phi} \theta_\text{new}
    \\ 
    & = \nabla_{f} \taskLoss \nabla_{\theta_\text{new}} f_{\theta_\text{new}} \nabla_{\phi} [\theta - \alpha \nabla_{\theta}  \mathbb{E}\left[\mathcal{M}_\phi(y, \ftheta{x})\right]
\vspace{-3pt}
\end{align}
where we first apply the chain rule and show that the gradient with respect to the meta-loss parameters $\phi$ requires the new model parameters $\theta_\text{new}$. 
We expand $\theta_\text{new}$ as one gradient step on $\theta$ based on meta-loss $\mathcal{M}_\phi$, making the dependence on $\phi$ explicit.

Optimization of the loss-parameters can either happen after \emph{each inner gradient} step (where inner refers to using the current loss parameters to update $\theta$), or after $M$ \emph{inner gradient steps} with the current meta-loss network $\mathcal{M}_\phi$.

The latter option requires back-propagation through a chain of all optimizee update steps. In practice we notice that updating the meta-parameters $\phi$ after each inner gradient update step works better. We reset $\theta$ after $M$ \emph{inner gradient steps}. We summarize the \emph{meta-train} phase in Algorithm~\ref{algo:ml3_train_sup}, with one \emph{inner} gradient step.

\vspace{-0.4cm}
\begin{figure}[!t]
\begin{minipage}[c]{0.49\textwidth}
\centering
\begin{algorithm}[H]
\begin{algorithmic}[1]
\footnotesize{
\STATE{$\phi \gets$ randomly initialize}
\WHILE{not done}
\STATE{$\theta \gets$ randomly initialize}
\STATE{$x, y \gets \text{Sample task samples from } \mathcal{T}$}
\STATE{$\learnedLoss = \mathcal{M}(y, \ftheta{x})$}
\STATE{$\theta_\text{new} \gets \theta - \alpha \nabla_{\theta} \mathop{\mathbb{E}}_{x}\left[\learnedLoss \right]$}
\STATE{$\phi \gets \phi - \eta \nabla_\phi \taskLoss(y, f_{\theta_\text{new}})$}
\ENDWHILE
}
\end{algorithmic}
\caption{ML$^3$ at (\textit{meta-train})}
\label{algo:ml3_train_sup}
\end{algorithm}

\vspace{-0.8cm}

\begin{algorithm}[H]
\begin{algorithmic}[1]

\footnotesize{
\STATE{$M \gets \text{\# of optimizee updates}$}
\STATE{$\theta \gets $ randomly initialize}
\FOR{$j \in \{0,\dots,M\}$}
\STATE{$x, y \gets \text{Sample task samples from } \mathcal{T}$}
\STATE{$\learnedLoss = \mathcal{M}(y, \ftheta{x})$}
\STATE{$\theta \gets \theta - \alpha \nabla_{\theta} \mathop{\mathbb{E}}_{x}\left[\learnedLoss \right]$}
\ENDFOR
}
\end{algorithmic}
\caption{ML$^3$ at (\textit{meta-test})}
\label{algo:ml3_test_sup}
\end{algorithm}
\end{minipage}
\vspace{-0.5cm}
\end{figure}

\vspace{5pt}
\subsection{\texorpdfstring{ML$^3$}{Lg} Reinforcement Learning}\label{sec:ml3_rl}
\vspace{-3pt}
In this section, we introduce several modifications that allow us to apply the ML$^3$ framework to reinforcement learning problems. 
Let $\mathcal{M} = (S, A, P, R, p_0, \gamma, T)$ be a finite-horizon Markov Decision Process (MDP), where $S$ and $A$ are state and action spaces, 
$P: S \times A \times S \rightarrow \mathbb{R}_{+}$ is a state-transition probability function or system dynamics, 
$R: S \times A \rightarrow \mathbb{R}$ a reward function, 
$p_0: S \rightarrow \mathbb{R}_{+}$ an initial state distribution, 
$\gamma$ a reward discount factor, and $T$ a horizon. 
Let $\tau = (s_0, a_0, \dots, s_T, a_T)$ be a trajectory of states and actions and 
$R(\tau) = \sum_{t=0}^{T-1} \gamma^t R(s_t,a_t)$ the trajectory return. 
The goal of reinforcement learning is to find parameters $\theta$ of a policy $\pi_\theta(a | s)$ that maximizes the expected discounted reward over trajectories induced by the policy: $\mathbb{E}_{\pi_\theta}[R(\tau)]$ where $ s_0\sim p_0, s_{t+1}\sim P(s_{t+1} | s_t, a_t)$ and $a_t\sim \pi_\theta(a_t | s_t)$. 
In what follows, we show how to train a meta-loss network to perform effective policy updates in a reinforcement learning scenario. 
To apply our ML$^3$ framework, we replace the optimizee $f_\theta$ from the previous section with a stochastic policy $\pi_\theta(a | s)$. 
We present two applications of ML$^3$ to RL. 

\subsubsection{\texorpdfstring{ML$^3$}{Lg} for Model-Based Reinforcement Learning}
\label{sec:ml3_mbrl}
Model-based RL (MBRL) attempts to learn a policy $\pi_\theta$ by first learning a dynamic model $P$. 
Intuitively, if the model $P$ is accurate, we can use it to optimize the policy parameters $\theta$. 
As we typically do not know the dynamics model a-priori, MBRL algorithms iterate between using the current approximate dynamics model $P$, to optimize the policy $\pi_\theta$ such that it maximizes the reward $R$ under $P$, then use the optimized policy $\pi_\theta$ to collect more data which is used to update the model $P$. 
In this context, we aim to learn a loss function that is used to optimize policy parameters through our meta-network $\mathcal{M}$.

Similar to the supervised learning setting we use current meta-parameters $\phi$ to optimize policy parameters $\theta$ under the current dynamics model $P$: 
$\theta_\text{new} = \theta - \alpha \nabla_{\theta} \left[\mathcal{M}_\phi(\tau, g)\right]$,

where $\tau = (s_0, a_0, \dots, s_T, a_T)$ is the sampled trajectory and the variable $g$ captures some task-specific information, such as the goal state of the agent.
To optimize $\phi$ we again need to define a task loss, which in the MBRL setting can be defined as $\mathcal{L}_{\mathcal{T}}(g, \pi_{\theta_\text{new}}) = -\mathbb{E}_{\pi_{\theta_\text{new}},P}[R_g(\tau_\text{new})]$,
denoting the reward that is achieved under the current dynamics model $P$. To update $\phi$, we compute the gradient of the task loss $\mathcal{L}_\mathcal{T}$ wrt. $\phi$, which involves differentiating all the way through the reward function, dynamics model and the policy that was updated using the meta-loss $\mathcal{M}_\phi$. 
The pseudo-code in Algorithm \ref{algo:ml3_mbrl}~(Appendix~\ref{apdx:algos}) illustrates the MBRL learning loop.  
In Algorithm \ref{algo:ml3_mbrl_test}~(Appendix~\ref{apdx:algos}), we show the policy optimization procedure during meta-test time. Notably, we have found that in practice, the model of the dynamics $P$ is not needed anymore for policy optimization at meta-test time. 
The meta-network learns to implicitly represent the gradients of the dynamics model and can produce a loss to optimize the policy directly. 

\subsubsection{\texorpdfstring{ML$^3$}{Lg} for Model-Free Reinforcement Learning}

Finally, we consider the model-free reinforcement learning (MFRL) case, where we learn a policy without learning a dynamics model. 
In this case, we can define a surrogate objective, which is independent of the dynamics model, as our task-specific loss~\citep{reinforce,sutton00,SchulmanHWA15}:
\vspace{-3.5pt}
\begin{align}
\label{eq:mfrl_taskloss}
    \mathcal{L}_{\mathcal{T}}(g, \pi_{\theta_\text{new}}) = -\mathbb{E}_{\pi_{\theta_\text{new}}}\left[R_g(\tau_\text{new})\log \pi_{\theta_\text{new}}(\tau_\text{new})\right] \\ =  -\mathbb{E}_{\pi_{\theta_\text{new}}}\left[R_g(\tau_\text{new}) \sum_{t=0}^{T-1} \log \pi_{\theta_\text{new}}(a_t | s_t)\right]
\\[-17.6pt]
\notag
\end{align}
Similar to the MBRL case, the task loss is indirectly a function of the meta-parameters $\phi$ that are used to update the policy parameters. 
Although we are evaluating the task loss on full trajectory rewards, we perform policy updates from Eq.~\ref{eq:pol_update} using stochastic gradient descent (SGD) on the meta-loss with mini-batches of experience $\left(s_i, a_i, r_i\right)$ for $i \in \{0,\dots,B-1\}$ with batch size $B$, similar to~\citet{epg18}. 
The inputs of the meta-loss network are the sampled states, sampled actions, task information $g$ and policy probabilities of the sampled actions: $\mathcal{M}_\phi\left(s, a, \pi_\theta(a | s), g\right)$. 
In this way, we enable efficient optimization of very high-dimensional policies with SGD provided only with trajectory-based rewards.
In contrast to the above MBRL setting, the rollouts used for task-loss evaluation are real system rollouts, instead of simulated rollouts. 
At test time, we use the same policy update procedure as in the MBRL setting, see Algorithm~\ref{algo:ml3_mbrl_test}~(Appendix~\ref{apdx:algos}).
\subsection{Shaping \texorpdfstring{ML$^3$}{Lg} loss by adding extra loss information during \emph{meta-train}}
\label{sec:extra_info}

So far, we have discussed using standard task losses, such as MSE-loss for regression or reward functions for RL settings. However, it is possible to provide more information about the task at \emph{meta-train} time, which can influence the learning of the loss-landscape. We can design our task-losses to incorporate extra penalties; for instance we can extend the MSE-loss with $\mathcal{L}_\text{extra}$ and weight the terms with $\beta$ and $\gamma$:
\begin{align}
    \mathcal{L}_\mathcal{T} = \beta(y - f_\theta(x))^2 + \gamma\mathcal{L}_\text{extra}
\end{align}
In our work, we experiment with 4 different types of extra loss information at \emph{meta-train} time: for supervised learning we show that adding extra information through $\mathcal{L}_\text{extra} = (\theta - \theta^*)^2$, where $\theta^*$ are the optimal regression parameters, can help shape a convex loss-landscape for otherwise non-convex optimization problems; we also show how we can use $\mathcal{L}_\text{extra}$ to induce a physics prior in robot model learning. For reinforcement learning tasks we demonstrate that by providing additional rewards in the task loss during meta-train time, we can encourage the trained meta-loss to learn exploratory behaviors; and finally also for reinforcement learning tasks, we show how expert demonstrations can be incorporated to learn loss functions which can generalize to new tasks. In all settings, the additional information shapes the learned loss function such that the environment does not need to provide this information during meta-test time.

\section{Experiments}
In this section we evaluate the applicability and the benefits of the learned meta-loss from two different view points. First, we study the benefits of using standard task losses, such as the mean-squared error loss for regression, to train the meta-loss in Section~\ref{sec:mimic_exp}. 
We analyze how a learned meta-loss compares to using a standard task-loss in terms of generalization properties and convergence speed. 
Second, we study the benefit of adding extra information at \emph{meta-train} time to shape the loss landscape in Section~\ref{sec:learned_landscape}. 
\subsection{Learning to mimic and improve over known task losses}\label{sec:mimic_exp}
First, we analyze how well our meta-learning framework can learn to mimic and improve over standard task losses for both supervised and reinforcement learning settings. For these experiments, the meta-network is parameterized by a neural network with two hidden layers of 40 neurons each. 
\subsubsection{Meta-Loss for Supervised Learning}
\begin{figure*}[t]
\centering

\begin{subfigure}[b]{0.24\textwidth}
\includegraphics[width=\textwidth]{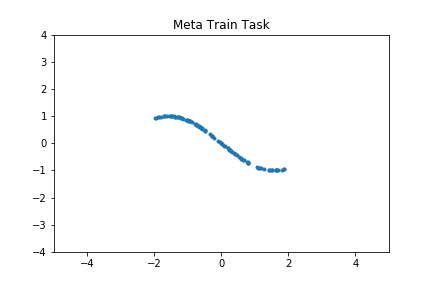}
\label{fig:meta_train_data_sine}
\end{subfigure}
\begin{subfigure}[b]{0.24\textwidth}
\includegraphics[width=\textwidth]{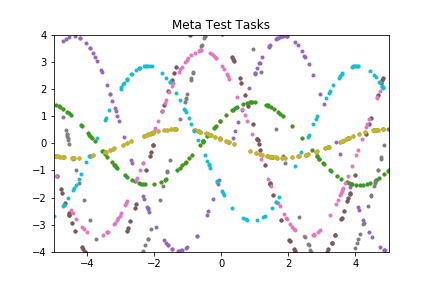}
\label{fig:meta_test_data_sine}
\end{subfigure}
\begin{subfigure}[b]{0.24\textwidth}
\includegraphics[width=\textwidth]{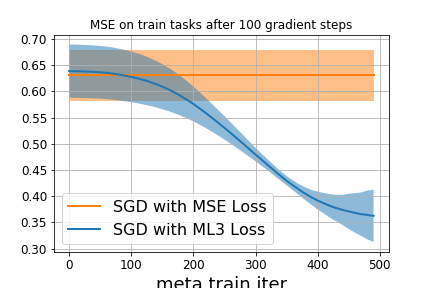}
\label{fig:meta_train_accuracy_sine}
\end{subfigure}
\begin{subfigure}[b]{0.24\textwidth}
\includegraphics[width=\textwidth]{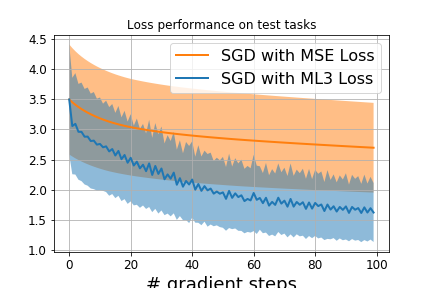}
\vspace{0.04cm}
\label{fig:meta_test_accuracy_sine}
\end{subfigure}
\vspace{-0.2cm}
\begin{subfigure}[b]{0.24\textwidth}
\includegraphics[width=\textwidth]{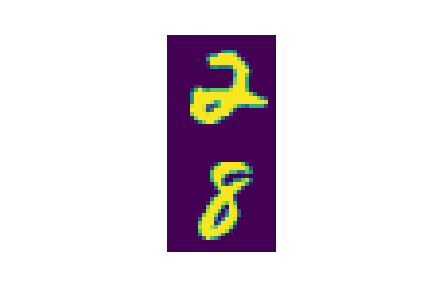}
\subcaption{Meta-Train Tasks}
\label{fig:meta_train_data_mnist}
\end{subfigure}
\begin{subfigure}[b]{0.24\textwidth}
\includegraphics[width=\textwidth]{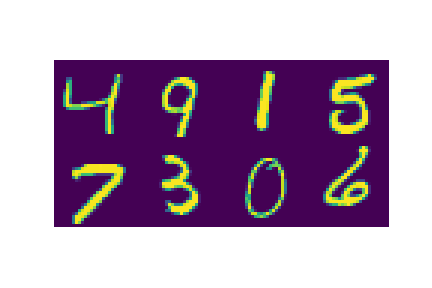}
\subcaption{Meta-Test Tasks}
\label{fig:meta_test_data_mnist}
\end{subfigure}
\begin{subfigure}[b]{0.24\textwidth}
\includegraphics[width=\textwidth, trim={0 0 0 0pt},clip]{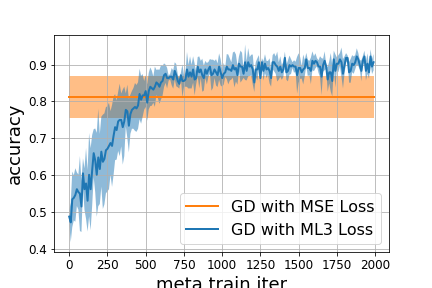}
\subcaption{Meta-Train}
\label{fig:meta_train_accuracy_mnist}
\end{subfigure}
\begin{subfigure}[b]{0.24\textwidth}
\includegraphics[width=\textwidth, trim={0 0 0 0pt},clip]{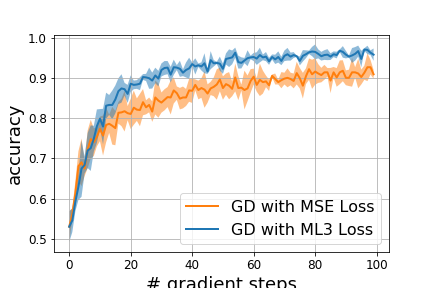}
\subcaption{Meta-Test}
\label{fig:meta_test_accuracy_mnist}
\end{subfigure}
\caption{\small Meta-learning for regression (top) and binary classification (bottom) tasks. 
(a) meta-train task, (b) meta-test tasks, (c) performance of the meta-network on the meta-train task as a function of (outer) meta-train iterations in blue, as compared to SGD using the task-loss directly in orange, (d) average performance of meta-loss on meta-test tasks as a function of the number of gradient update steps}
\label{fig:sup_exp}
\vspace{-0.5cm}
\end{figure*}

In this set of experiments, we evaluate how well our meta-learning framework can learn loss functions $\mathcal{M}_\phi$ for regression and classification tasks. 
In particular, we perform experiments on sine function regression and binary classification of digits (see details in Appendix~\ref{app:sine_task_details}). 
At meta-train time, we randomly draw one task for meta-training (see~\cref{fig:sup_exp} (a)), and at meta-test time we randomly draw $10$ test tasks for regression, and $4$ test tasks for classification (\cref{fig:sup_exp}(b)). 
For the sine regression, tasks are drawn according to details in Appendix~\ref{app:sine_task_details}, and we initialize our model $f_\theta$ to a simple feedforward NN with 2 hidden layers and 40 hidden units each, for the binary classification task $f_\theta$ is initialized via the \emph{LeNet} architecture~\citep{lenet}. 
For both experiments we use a fixed learning rate $\alpha=\eta=0.001$ for both inner ($\alpha$) and outer ($\eta$) gradient update steps. 
We average results across $5$ random seeds, where each seed controls the initialization of both initial model and meta-network parameters, as well as the the random choice of meta-train/test task(s), and visualize them in~\cref{fig:sup_exp}.
We compare the performance of using SGD with the task-loss $\mathcal{L}$ directly (in orange) to SGD using the learned meta-network $\mathcal{M}_\phi$ (in blue), both using a learning rate $\alpha=0.001$. 
In~\cref{fig:sup_exp} (c) we show the average performance of the meta-network $\mathcal{M}_\phi$ as it is being learned, as a function of (outer) meta-train iterations in blue. 
In both regression and classification tasks, the meta-loss eventually leads to a better performance on the meta-train task as compared to the task loss. In~\cref{fig:sup_exp} (d) we evaluate SGD using $\mathcal{M}_\phi$ vs SGD using $\mathcal{L}$ on previously unseen (and out-of-distribution) meta-test tasks as a function of the number of gradient steps. 
Even on these novel test tasks, our learned $\mathcal{M}_\phi$ leads to improved performance as compared to the task-loss.
\subsubsection{Learning Reward functions for Model-based Reinforcement Learning}
\begin{figure*}[t]
\centering
\begin{subfigure}[b]{0.3\textwidth}
\center
\includegraphics[width=0.95\textwidth]{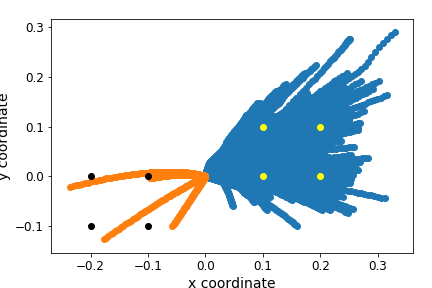}
\vspace{0.15cm}
\caption{train (blue), test (orange) tasks}
\label{fig:pointmass_model_based}
\end{subfigure}
\begin{subfigure}[b]{0.3\textwidth}
\center
\includegraphics[width=\textwidth]{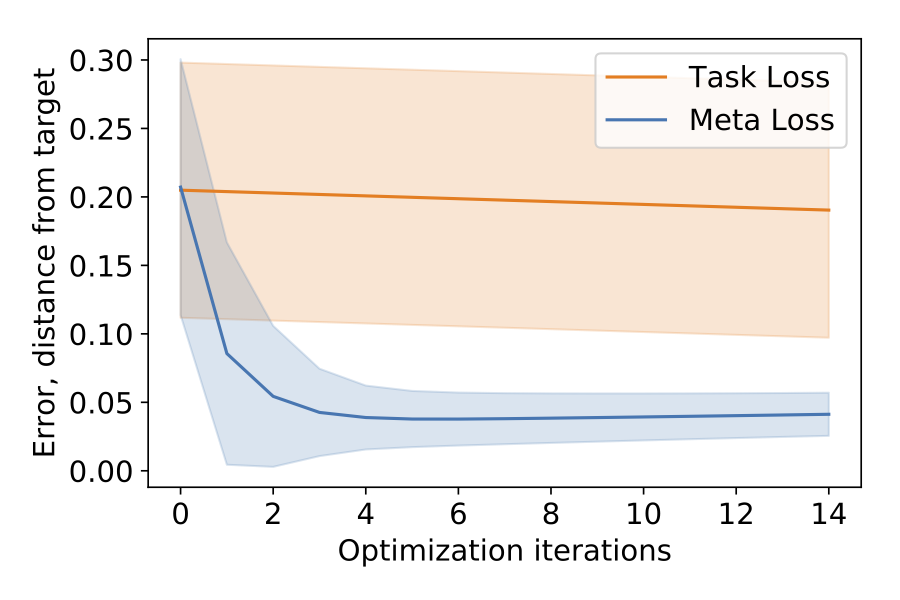}
\caption{Meta vs Task Loss Pointmass}
\label{fig:pointmass_model_based_losses}
\end{subfigure}
\begin{subfigure}[b]{0.3\textwidth}
\center
\includegraphics[width=\textwidth]{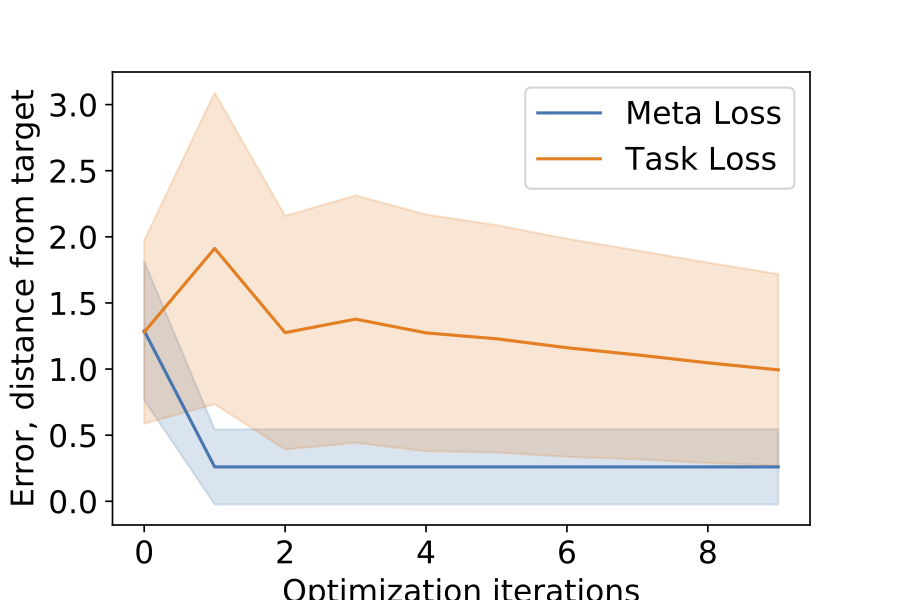}
\vspace{-7pt}
\caption{Meta vs Task Loss Reacher}
\label{fig:reacher_model_based_losses}
\end{subfigure}
\caption{ML$^{3}$ for MBRL: results are averaged across 10 runs. 
We can see in (a) that the ML$^{3}$ loss generalizes well, the loss was trained on the blue trajectories and tested on the orange ones for the PointmassGoal task. 
ML$^3$ loss also significantly speeds up learning when compared to the task loss at meta-test time on the PointmassGoal (b) and the ReacherGoal (c) environments.
}
\label{fig:model_based}
\end{figure*}

In the MBRL example, the tasks consist of a free movement task of a point mass in a 2D space, we call this environment PointmassGoal, and a reaching task with a 2-link 2D manipulator, which we call the ReacherGoal environment (see Appendix~\ref{app:mbrl_details} for details). 
The task distribution $p(\mathcal{T})$ consists of different target positions that either the point mass or the arm should reach. During meta-train time, a model of the system dynamics, represented by a neural network, is learned from samples of
the currently optimal policy. 
The task loss during meta-train time is $\mathcal{L}_{\mathcal{T}}(\theta) = \mathbb{E}_{\pi_{\theta},P}[R(\tau)]$, where $R(\tau)$ is the final distance from the goal $g$, when rolling out $\pi_{\theta_\text{new}}$ in the dynamics model $P$. Taking the gradient $\nabla_\phi \mathbb{E}_{\pi_{\theta_\text{new},P}}[R(\tau)]$ requires the differentiation through the learned model $P$ (see Appendix~\ref{algo:ml3_mbrl}). 
The input to the meta-network is the state-action trajectory of the current roll-out and the desired target position. The meta-network outputs a loss signal together with the learning rate to optimize the policy. \cref{fig:pointmass_model_based} shows the qualitative reaching performance of a policy optimized with the meta loss during meta-test on PointmassGoal. 
The meta-loss network was trained only on tasks in the right quadrant (blue trajectories) and tested on the tasks in the left quadrant (orange trajectories) of the $x,y$ plane, showing the generalization capability of the meta loss. Figure~\ref{fig:pointmass_model_based_losses} and \ref{fig:reacher_model_based_losses} show a comparison in terms of final distance to the target position at test time. The performance of policies trained with the meta-loss is compared to policies trained with the task loss, in this case final distance to the target. The curves show results for 10 different goal positions (including goal positions where the meta-loss needs to generalize). 
When optimizing with the task loss, we use the dynamics model learned during the meta-train time, as in this case the differentiation through the model is required during test time. As mentioned in Section~\ref{sec:ml3_mbrl}, this is not needed when using the meta-loss. 
\subsubsection{Learning Reward functions for Model-free Reinforcement Learning}
\begin{figure*}[ht]
\centering
\begin{subfigure}[b]{0.24\textwidth}
\includegraphics[width=0.9\textwidth]{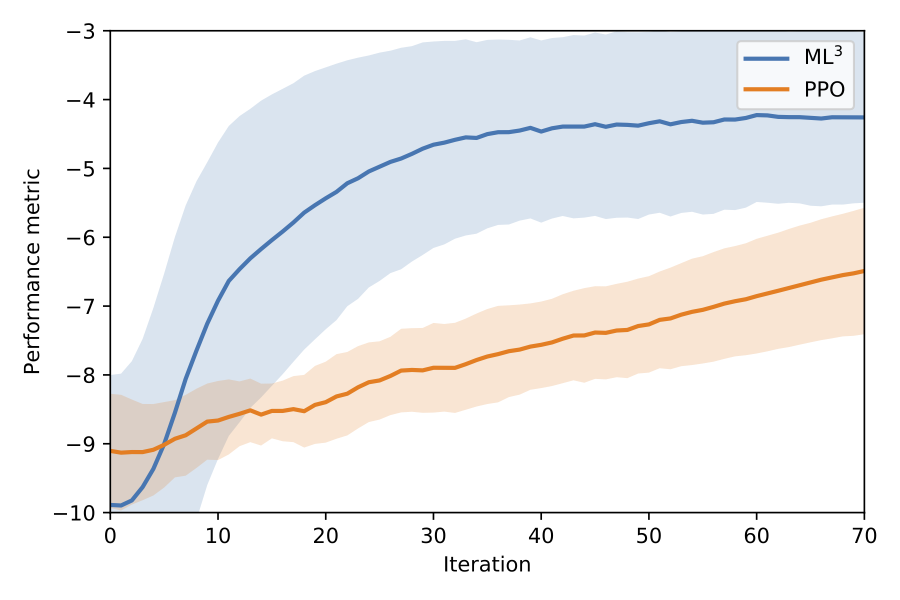}
\caption{ReacherGoal}\label{fig:reacher}
\end{subfigure}
\begin{subfigure}[b]{0.24\textwidth}
\includegraphics[width=0.9\textwidth]{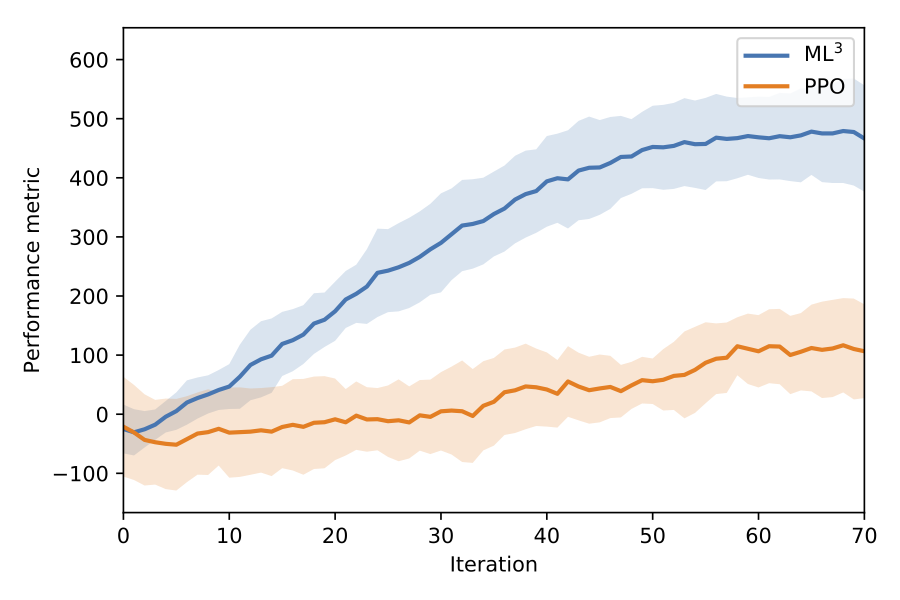}
\caption{AntGoal}\label{fig:Ant}
\end{subfigure}
\begin{subfigure}[b]{0.24\textwidth}
\includegraphics[width=0.9\textwidth]{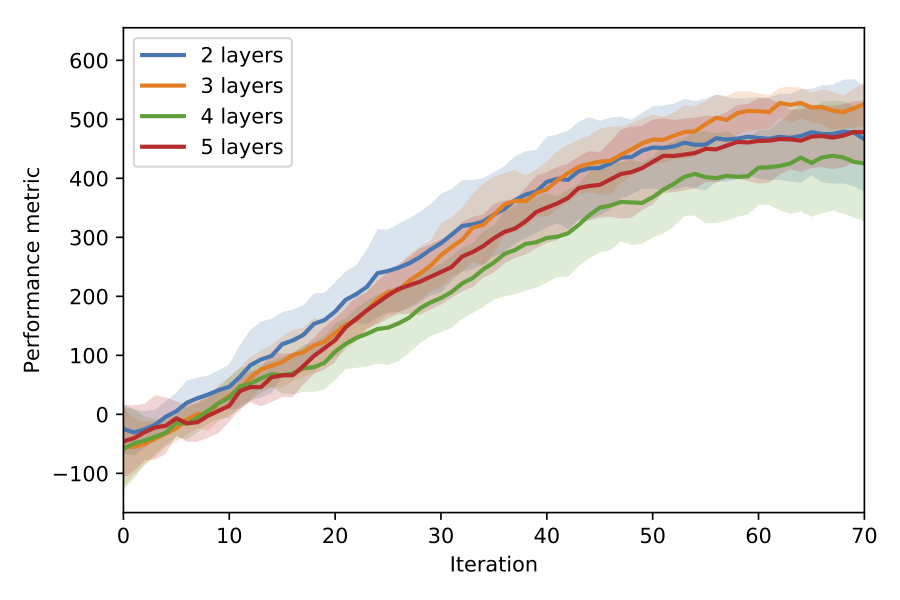}
\caption{ReacherGoal}\label{fig:reacher_architectures}
\end{subfigure}
\begin{subfigure}[b]{0.24\textwidth}
\includegraphics[width=0.9\textwidth]{fig/ml3_antgoal_architectures.png}
\caption{AntGoal}\label{fig:ant_architectures}
\end{subfigure}
\vspace{-6pt}
\caption{\small ML$^{3}$ for model-free RL: results are averaged across $10$ tasks. (a+b) Policy learning on new task with ML$^3$ loss compared to PPO objective performance during \emph{meta-test} time. The learned loss leads to faster learning at meta-test time. (c+d) Using the same ML$^3$ loss, we can optimize policies of different architectures, showing that our learned loss maintains generality.}
\label{fig:model_free}
\end{figure*}

In the following, we move to evaluating on model-free RL tasks. 
\cref{fig:model_free} shows results when using two continuous control tasks based on  OpenAI Gym MuJoCo environments~\citep{mujoco}: ReacherGoal and AntGoal (see Appendix~\ref{app:mfrl_details} for details)\footnote{Our framework is implemented using open-source libraries \textit{Higher}~\cite{grefenstette2019generalized} for convenient second-order derivative computations and \textit{Hydra}~\citep{hydra2020} for simplified handling of experiment configurations.}
\cref{fig:reacher} and \cref{fig:Ant} show the results of the meta-test time performance for the ReacherGoal and the AntGoal environments respectively.
We can see that ML$^3$ loss significantly improves optimization speed in both scenarios compared to PPO.
In our experiments, we observed that on average ML$^{3}$ requires 5 times fewer samples to reach 80\% of task performance in terms of our metrics for the model-free tasks.

To test the capability of the meta-loss to generalize across different architectures, we first meta-train $\mathcal{M}_\phi$ on an architecture with two layers and meta-test the same meta-loss on architectures with varied number of layers.
\cref{fig:model_free}~(c+d) show meta-test time comparison for the ReacherGoal and the AntGoal environments in a model-free setting for four different model architectures.
Each curve shows the average and the standard deviation over ten different tasks in each environment.
Our comparison clearly indicates that the meta-loss can be effectively re-used across multiple architectures with a mild variation in performance compare to the overall variance of the corresponding task optimization.

\subsection{Shaping loss landscapes by adding extra information at meta-train time}\label{sec:learned_landscape}
This set of experiments shows that our meta-learner is able to learn loss functions that incorporate extra information available only during meta-train time. The learned loss will be shaped such that optimization is faster when using the meta-loss compared to using a standard loss.
\subsubsection{Illustration: Shaping loss}
We start by illustrating the loss shaping on an example of sine frequency regression where we fit a single parameter for the purpose of visualization simplicity. 

\begin{figure*}[ht]
\centering
\vspace{-7pt}
\begin{subfigure}[b]{0.24\textwidth}
\includegraphics[width=\textwidth]{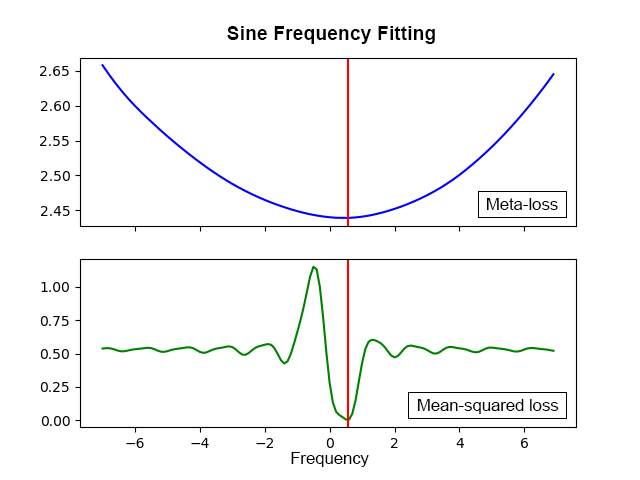}
\caption{{\small Sine: learned vs task loss}}
\label{fig:sine_learned_vs_task_loss}
\end{subfigure}
\begin{subfigure}[b]{0.24\textwidth}
\includegraphics[width=\textwidth]{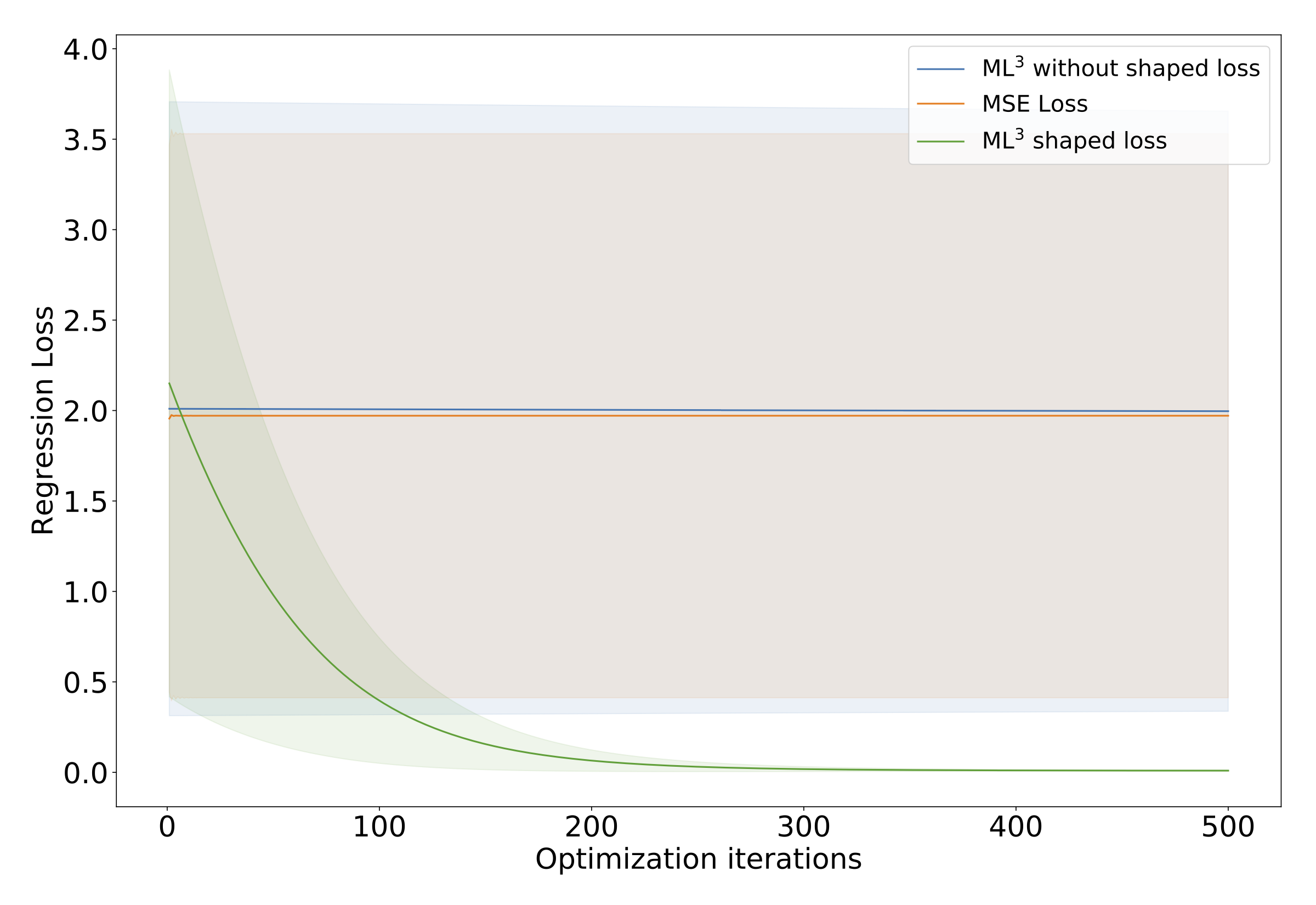}
\caption{{\small Sine: meta-test time}}
\label{fig:meta_test_sine}
\end{subfigure}
\begin{subfigure}[b]{0.25\textwidth}
\includegraphics[width=\textwidth]{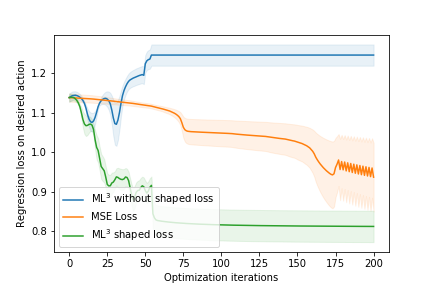}
\caption{Reacher: inverse dynamics}\label{fig:meta_inverse_dynamics_learning_reacher}
\end{subfigure}
\begin{subfigure}[b]{0.25\textwidth}
\includegraphics[width=\textwidth]{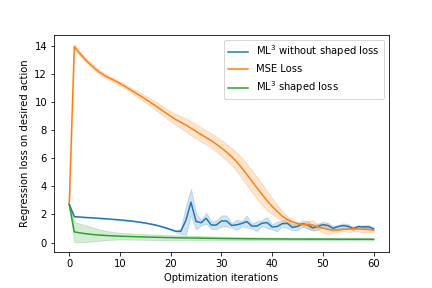}
\caption{Sawyer: inverse dynamics}\label{fig:meta_inverse_dynamics_learning_7dof}
\end{subfigure}
\vspace{-5pt}
\caption{
\small Meta-test time evaluation of the shaped meta-loss (ML$^3$), i.e. trained with shaping ground-truth (extra) information at meta-train time:
a) Comparison of learned ML$^3$ loss (top) and MSE loss (bottom) landscapes for fitting the frequency of a sine function. The red lines indicate the ground-truth values of the frequency. 
b) Comparing optimization performance of: ML$^3$ loss trained with (green), and without (blue) ground-truth frequency values; MSE loss (orange). The ML$^3$ loss learned with the ground-truth values outperforms both the non-shaped ML$^3$ loss and the MSE loss. %
c-d) Comparing performance of inverse dynamics model learning for ReacherGoal (c) and Sawyer arm (d). ML$^3$ loss trained with (green) and without (blue) ground-truth inertia matrix is compared to MSE loss (orange). The shaped ML$^3$ loss outperforms the  MSE loss in all cases.
}
\label{fig:sine_inv_dyn}
\end{figure*}

For this illustration we generate training data \mbox{$\mathcal{D} =\{x_n, y_n\}^N, N=1000$}, by drawing data samples from the ground truth function $y = sin(\nu x)$, for $x= [-1, 1]$. 
We create a model $f_{\omega}(x) = \sin(\omega x)$, and aim to optimize parameter $\omega$ on $\mathcal{D}$, with the goal of recovering value $\nu$. 
\cref{fig:sine_learned_vs_task_loss} (bottom) shows the loss landscape for optimizing $\omega$, when using the MSE loss. 
The target frequency $\nu$ is indicated by a vertical red line. 
As noted by~\citet{sines}, the landscape of this loss is highly non-convex and difficult to optimize with conventional gradient descent.

Here, we show that by utilizing additional information about the ground truth value of the frequency at meta-train time, we can learn a better shaped loss.
Specifically, during meta-train time, our task-specific loss is the squared distance to the ground truth frequency: $(\omega - \nu)^2$ that we later call the \textit{shaping loss}. 
The inputs of the meta-network $\mathcal{M}_\phi(y, \hat{y})$ are the training targets $y$ and predicted  function values $\hat{y} = f_\omega(x)$, similar to the inputs to the mean-squared loss. 
After meta-train time commences our learned loss function $\mathcal{M}_\phi$ produces a convex loss landscapes as depicted in \cref{fig:sine_learned_vs_task_loss}(top). 

To analyze how the shaping loss impacts model optimization at meta-test time, we compare 3 loss functions: 1) directly using standard MSE loss (orange), 2) ML$^3$ loss that was trained via the MSE loss as task loss (blue), and 3) ML$^3$ loss trained via the shaping loss, \cref{fig:meta_test_sine}. When comparing the performance of these 3 losses, it becomes evident that without shaping the loss landscape, the optimization is prone to getting stuck in a local optimum.

\begin{figure*}[ht]
    \centering
    \begin{subfigure}[b]{0.25\textwidth}
\includegraphics[width=\textwidth]{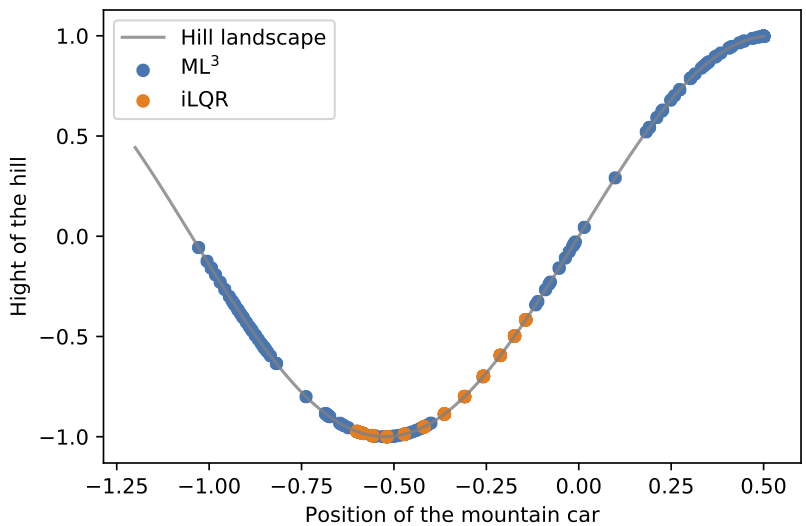}
\caption{Trajectory ML$^3$ vs. iLQR}
\end{subfigure}
\begin{subfigure}[b]{0.24\textwidth}
\includegraphics[width=\textwidth]{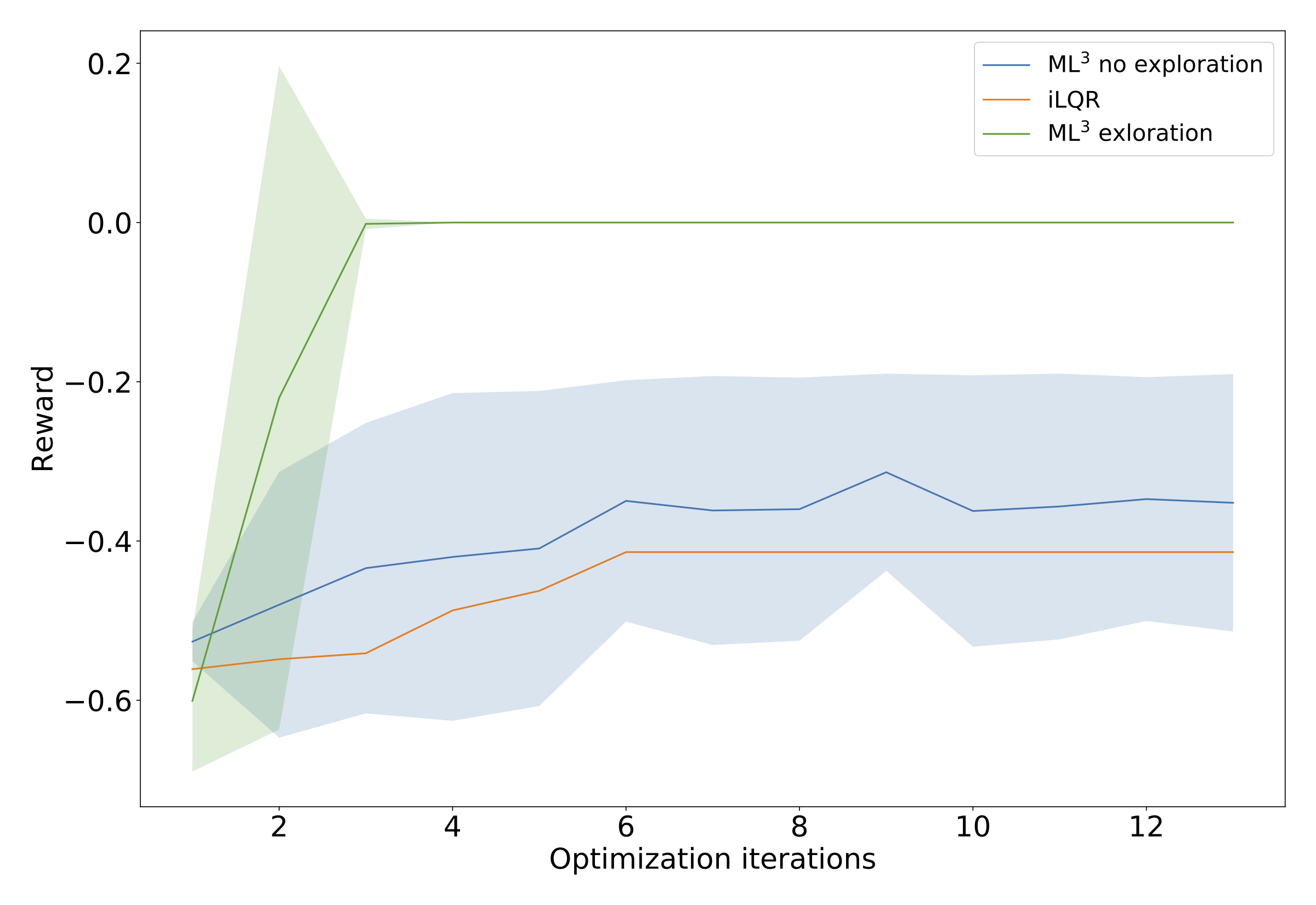}
\caption{MountainCar: meta-test time}
\end{subfigure}
\begin{subfigure}[b]{0.24\textwidth}
        \includegraphics[width=\textwidth,trim={0 0 0 15pt},clip]{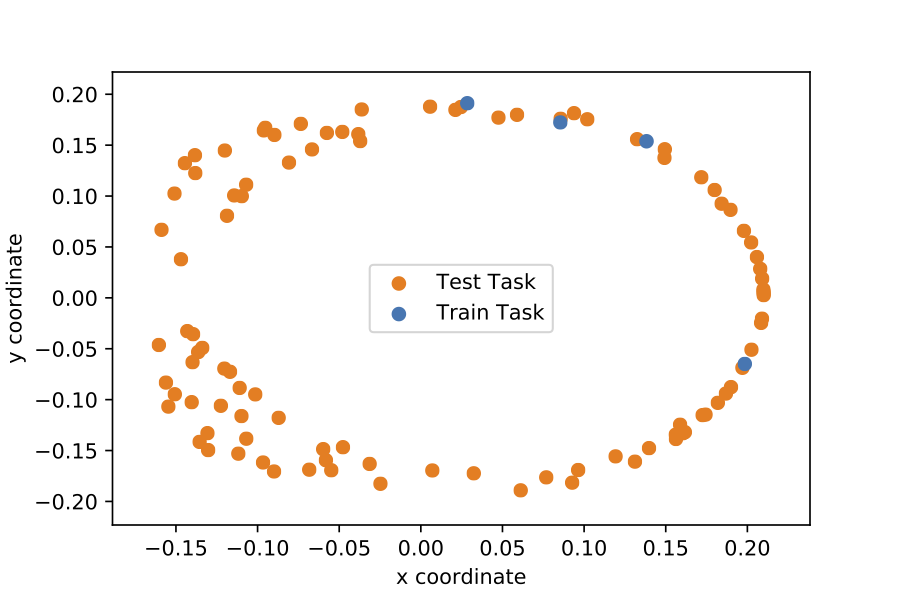}
        \vspace{-6pt}
        \caption{Train and test targets}
        \label{fig:reacher_bc_experiment_tasks}
\end{subfigure}
\begin{subfigure}[b]{0.24\textwidth}
        \includegraphics[width=\textwidth,trim={0 0 0 15pt},clip]{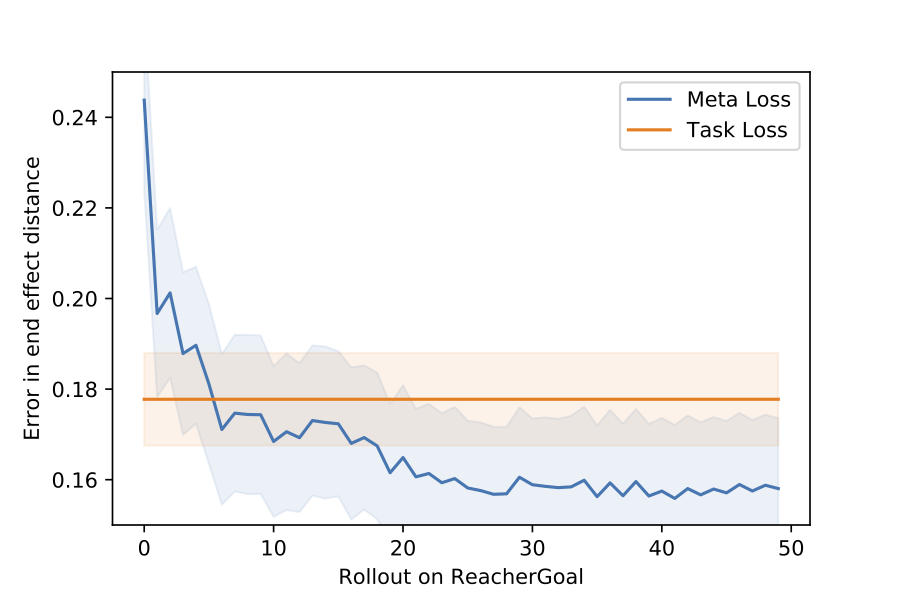}
        \vspace{-6pt}
        \caption{ML$^3$ vs. Task loss at test}
        \label{fig:reacher_bc_experiment}
\end{subfigure}
\vspace{-5pt}
\caption{(a) MountainCar trajectory for policy optimized with iLQR compared to ML$^3$ loss with extra information. (b) optimization performance during meta-test time for policies optimized with iLQR compared to ML$^3$ with and without extra information. (c+d) ReacherGoal with expert demonstrations available during meta-train time. (c) shows the targets in end-effector space. The four blue dots show the training targets for which expert demonstrations are available, the orange dots show the meta-test targets. In (d) we show the reaching performance of a policy trained with the shaped ML$^3$ loss at meta-test time, compared to the performance of training simply on the behavioral cloning objective and testing on test targets.}
    \label{fig:intermediate_goals_and_expert_demos}
\end{figure*}
\subsubsection{Shaping loss via physics prior for inverse dynamics learning}
Next, we show the benefits of shaping our ML$^3$ loss via ground truth parameter information for a robotics application. Specifically, we aim to learn and shape a meta-loss that improves sample efficiency for learning (inverse) dynamics models, i.e. 
a mapping $u=f(q, \dot{q}, \ddot{q}_\text{des})$, where: $q$, $\dot{q}$, $\ddot{q}_\text{des}$ are vectors of joint angular positions, velocities and desired accelerations; $u$ is a vector of joint torques.

Rigid body dynamics (RBD) provides an analytical solution to computing the (inverse) dynamics and can generally be written as:
\begin{equation}
M(q)\ddot{q} + F(q,\dot{q}) = u
\label{eq:equation_of_motion}  
\end{equation}
where the inertia matrix $M(q)$, and $F(q,\dot{q})$ are computed analytically~\cite{featherstone2014rigid}. 
Learning an inverse dynamics model using neural networks can increase the expressiveness compared to RBD but requires many data samples that are expensive to collect.
Here we follow the approach in~\citep{lutter2019deep}, and attempt to learn the inverse dynamics via a neural network that predicts the inertia matrix $M_\theta(q)$. To improve upon sample efficiency we apply our method by shaping the loss landscape during meta-train time using the ground truth inertia matrix $M(q)$ provided by a simulator. Specifically, we use the task loss $\mathcal{L}_\mathcal{T} =(M_\theta(q)-M(q))^2$ to optimize our meta-loss network. During meta-test time we use our trained meta-loss shaped with the physics prior (the inertia matrix exposed by the simulator) to optimize the inverse dynamics neural network.
In~\cref{fig:sine_inv_dyn}-c we show the prediction performance of the inverse dynamics model during meta-test time on new trajectories of the ReacherGoal environment. 
We compare the optimization performance during meta-test time when using the meta-loss trained with physics prior, the meta loss trained without physics prior (i.e via MSE loss) to the optimization with MSE loss.
\cref{fig:sine_inv_dyn}-d shows a similar comparison for the Sawyer environment - a simulator of the 7 degrees-of-freedom Sawyer anthropomorphic robot arm. 
Inverse dynamics learning using the meta loss with physics prior achieves the best prediction performance on both robots. ML$^3$ without physics prior performs worst on the ReacherGoal environment, in this case the task loss formulated only in the action space did not provide enough information to learn a $\mathcal{L}_\text{learned}$ useful for optimization. For the Sawyer training with MSE loss leads to a slower optimization, however the asymptotic performance of MSE and ML$^3$ is the same. Only ML$^3$ with shaped loss outperforms both.
\subsubsection{Shaping Loss via intermediate goal states for RL}
We analyze loss landscape shaping on the MountainCar environment~\citep{Moore1990}, a classical control problem where an under-actuated car has to drive up a steep hill. 
The propulsion force generated by the car does not allow steady climbing of the hill, thus
greedy minimization of the distance to the goal often results in a failure to solve the task.
The state space is two-dimensional consisting of the  position and velocity of the car, the action space consists of a one-dimensional torque. In our experiments, we provide intermediate goal positions during meta-train time, which are not available during the meta-test time. 
The meta-network incorporates this behavior into its loss leading to an improved exploration during the meta-test time as can be seen in~\cref{fig:intermediate_goals_and_expert_demos}-a, when compared to a classical iLQR-based trajectory optimization~\citep{Tassa2014}. \cref{fig:intermediate_goals_and_expert_demos}-b~shows the average distance between the car and the goal at last rollout time step over several iterations of policy updates with ML$^3$ with and without extra information and iLQR. As we observe, ML$^3$ with extra information can successfully bring the car to the goal in a small amount of updates, whereas iLQR and ML$^3$ without extra information is not able to solve this task.

\subsubsection{Shaping loss via expert information during meta-train time}
Expert information, like demonstrations for a task, is another way of adding relevant information during meta-train time, and thus shaping the loss landscape. In \textit{learning from demonstration (LfD)}~\citep{pomerleau1991efficient, ng2000algorithms,BillardCDS08}, expert demonstrations are used for initializing robotic policies. In our experiments, we aim to mimic the availability of an expert at meta-test time by training our meta-network to optimize a behavioral cloning objective at meta-train time. We provide the meta-network with expert state-action trajectories during train time, which could be human demonstrations or, as in our experiments, trajectories optimized using iLQR. During meta-train time, the task loss is the behavioral cloning objective $\mathcal{L}_{\mathcal{T}}(\theta) =\mathbb{E}\left[\sum_{t=0}^{T-1}[\pi_{\theta_{\text{new}}}(a_t|s_t) - \pi_{\text{expert}}(a_t|s_t)]^2\right]$. Fig.~\ref{fig:reacher_bc_experiment} shows the results of our experiments in the ReacherGoal environment.

\section{Conclusions}
In this work we presented a framework to meta-learn a loss function entirely from data. 
We showed how the meta-learned loss can become well-conditioned and suitable for an efficient optimization with gradient descent.
When using the learned meta-loss we observe significant speed improvements in regression, classification and benchmark reinforcement learning tasks. 
Furthermore, we showed that by introducing additional guiding information during training time we can train our meta-loss to develop exploratory strategies that can significantly improve performance during the meta-test time. 

We believe that the ML$^3$ framework is a powerful tool to incorporate prior experience and transfer learning strategies to new tasks. 
In future work, we plan to look at combining multiple learned meta-loss functions in order to generalize over different families of tasks. 
We also plan to further develop the idea of introducing additional curiosity rewards during training time to improve the exploration strategies learned by the meta-loss. 

\fontsize{8pt}{8pt}\selectfont
\section*{Acknowledgment}
The authors thank the International Max Planck Research School for Intelligent Systems (IMPRS-IS) for supporting Sarah Bechtle. This work was in part supported by New  York  University, the European Union's Horizon 2020 research and innovation program (grant agreement 780684 and European Research Councils grant 637935) and the National Science Foundation (grants 1825993 and 2026479).
\renewcommand{\bibfont}{\small}
\setlength\bibsep{0.19\baselineskip}
\bibliography{main.bib}
\bibliographystyle{abbrvnat}

\appendix

\section{MFRL and MBRL algorithms details}
\label{apdx:algos}
\begin{figure}[htb]
\begin{minipage}{0.45\textwidth}
\vspace{-0.7cm}
\begin{algorithm}[H]
\begin{algorithmic}[1]
\STATE{$\phi, \gets$ randomly initialize parameters}
\STATE{Randomly initialize dynamics model $P$}
\WHILE{not done}
\STATE{$\theta \gets$ randomly initialize parameters}
\STATE{$\tau \gets \text{forward unroll } \pi_{\theta} \text{ using } P$}
\STATE{$\pi_{\theta_\text{new}} \gets \text{ optimize}(\tau, \mathcal{M}_\phi, g, R)$}
\STATE{$\tau_\text{new} \gets \text{forward unroll } \pi_{\theta_\text{new}} \text{ using } P$}
\STATE{\text{\emph{Update $\phi$ to maximize reward under $P$:}}}
 \STATE{$\phi \gets \phi - \eta \nabla_\phi \mathcal{L}_\mathcal{T}(\tau_\text{new})$}
\STATE{$\tau_{real} \gets \text{roll out } \pi_{\theta_\text{new}} \text{ on real system} $}
\STATE{$P \gets $ update dynamics model with $\tau_{real}$}
\ENDWHILE
\end{algorithmic}
\caption{ML$^3$ for MBRL \emph(meta-train)}
\label{algo:ml3_mbrl}
\end{algorithm}
\end{minipage}
%
\begin{minipage}{0.45\textwidth}
\begin{algorithm}[H]
\begin{algorithmic}[1]
\footnotesize{
\STATE{$I \gets$ \# of inner steps}
\STATE{$\phi \gets$ randomly initialize parameters}
\WHILE{not done}
\STATE{$\theta_\text{0} \gets$ randomly initialize policy}
\STATE{$\mathcal{T} \gets$ sample training tasks}
\STATE{$\tau_\text{0}, R_\text{0} \gets \text{roll out policy } \pi_{\theta_\text{0}}$}
\FOR{$i \in \{0,\dots,I\}$}
  \STATE{$\pi_{\theta_\text{i+1}} \gets \text{ optimize}(\pi_{\theta_\text{i}}, \mathcal{M}_\phi, \tau_{i}, R_{i})$}
  \STATE{$\tau_{i+1}, R_{i+1} \gets \text{roll out policy } \pi_{\theta_\text{i+1}}$}
  \STATE{$\mathcal{L}^{i}_\mathcal{T} \gets$ compute task-loss $\mathcal{L}^{i}_\mathcal{T}(\tau_{i+1}, R_{i+1})$}
\ENDFOR
\STATE{$\mathcal{L}_\mathcal{T} \gets \mathop{\mathbb{E}}\left[\mathcal{L}^{i}_\mathcal{T}\right]$}
\STATE{$\phi \gets \phi - \eta \nabla_\phi \mathcal{L}_\mathcal{T}$}
\ENDWHILE
}
\end{algorithmic}
\caption{ML$^3$ for MFRL (\textit{meta-train})}
\label{algo:ml3_train_rl}
\end{algorithm}
\vspace{-0.4cm}
\begin{algorithm}[H]
\begin{algorithmic}[1]
\footnotesize{
\STATE{$\theta \gets$ randomly initialize policy}
\FOR{$j \in \{0,\dots,M\}$}
\STATE{$\tau, R \gets \text{roll out } \pi_\theta $}
\STATE{$\pi_{\theta} \gets \text{ optimize}(\pi_{\theta}, \mathcal{M}_\phi, \tau, R)$}
\ENDFOR
}
\end{algorithmic}
\caption{ML$^3$ for RL \emph(meta-test)}
\label{algo:ml3_mbrl_test}
\end{algorithm}
\end{minipage}
\vspace{-5pt}
\end{figure}

We notice that in practice, including the policy's distribution parameters directly in the meta-loss inputs, e.g. mean $\mu$ and standard deviation $\sigma$ of a Gaussian policy, works better than including the probability estimate $\pi_\theta(a | s)$, as it provides a direct way to update the distribution parameters using back-propagation through the meta-loss.

\section{Experiments: MBRL}\label{app:mbrl_details}
The forward model of the dynamics is represented in both cases by a neural network, the input to the network is the current state and action, the output is the next state of the environment.

The Pointmass state space is four-dimensional. For PointmassGoal $(x,y,\dot{x},\dot{y})$ are the 2D positions and velocities, and the actions are accelerations $(\ddot{x}, \ddot{y})$.

The ReacherGoal environment for the MBRL experiments is a lower-dimensional variant of the MFRL environment. 
It has a four dimensional state, consisting of position and angular velocity of the joints $[\theta_1,\theta_2,\dot{\theta_1},\dot{\theta_2}]$ the torque is two dimensional $[\tau_1,\tau_2]$
The dynamics model $P$ is updated once every 100 outer iterations with the samples collected by the policy from the last inner optimization step of that outer optimization step, i.e. the latest policy.


\section{Experiments: MFRL}\label{app:mfrl_details}
The ReacherGoal environment is a 2-link 2D manipulator that has to reach a specified goal location with its end-effector.
The task distribution (at meta-train and meta-test time) consists of an initial link configuration and random goal locations within the reach of the manipulator.
The performance metric for this environment is the mean trajectory sum of negative distances to the goal, averaged over 10 tasks.
As a trajectory reward $R_g(\tau)$ for the task-loss (see Eq.~\ref{eq:mfrl_taskloss}) we use $R_g(\tau) = -d + 1/(d + 0.001) - |a_{t}|$
, where $d$ is the distance of the end-effector to the goal $g$ specified as a 2-d Cartesian position.
The environment has eleven dimensions specifying angles of each link, direction from the end-effector to the goal, Cartesian coordinates of the target and Cartesian velocities of the end-effector.

The AntGoal environment requires a four-legged agent to run to a goal location.
The task distribution consists of random goals initialized on a circle around the initial position.
The performance metric for this environment is the mean trajectory sum of differences between the initial and the current distances to the goal, averaged over 10 tasks. 
Similar to the previous environment we use $R_g(\tau) = -d + 5/(d + 0.25) - |a_{t}|$
, where $d$ is the distance from the center of the creature's torso to the goal $g$ specified as a 2D Cartesian position.
In contrast to the ReacherGoal this environment has $33$~\footnote{In contrast to the original Ant environment we remove external forces from the state.} dimensional state space that describes Cartesian position, velocity and orientation of the torso as well as angles and angular velocities of all eight joints. 
Note that in both environments, the meta-network receives the goal information $g$ as part of the state $s$ in the corresponding environments.
Also, in practice, including the policy's distribution parameters directly in the meta-loss inputs, e.g. mean $\mu$ and standard deviation $\sigma$ of a Gaussian policy, works better than including the probability estimate $\pi_\theta(a | s)$, as it provides a more direct way to update $\theta$ using back-propagation through the meta-loss.

\section{Experiments: Regression and Classification Details}\label{app:sine_task_details}
For the sine task at meta-train time, we draw $100$ data points from function $y = \sin{(x - \pi)}$, with $x \in [-2.0, 2.0]$.
For meta-test time we draw $100$ data points from function $y = A \sin{(x - \omega)}$, with $A \sim [0.2, 5.0]$, $\omega \sim [-\pi, pi]$ and $x \in [-2.0, 2.0]$.
We initialize our model $f_\theta$ to a simple feedforward NN with 2 hidden layers and 40 hidden units each, for the binary classification task $f_\theta$ is initialized via the \emph{LeNet} architecture.
For both regression and classification experiments we use a fixed learning rate $\alpha=\eta=0.001$ for both inner ($\alpha$) and outer ($\eta$) gradient update steps. We average results across $5$ random seeds, where each seed controls the initialization of both initial model and meta-network parameters, as well as the the random choice of meta-train/test task(s), and visualize them in~\cref{fig:sup_exp}.
Task losses are $\mathcal{L}_\text{Regression} = (y - f_\theta(x))^2$ and $\mathcal{L}_\text{BinClass} = \textit{CrossEntropyLoss}(y, f_\theta(x))$ for regression and classification meta-learning respectively.

\end{document}